\documentclass[10pt,twocolumn,letterpaper]{article}
\usepackage{iccv}
\usepackage{times}
\usepackage{epsfig}
\usepackage{graphicx}
\usepackage{amsmath}
\usepackage{multirow}
\usepackage{subfigure}
\usepackage{amssymb}
\usepackage{float}
\usepackage{booktabs}

\usepackage[pagebackref=true,breaklinks=true,letterpaper=true,colorlinks,bookmarks=false]{hyperref}

\iccvfinalcopy 

\def\iccvPaperID{****} 

\ificcvfinal\fi

\begin{document}

\title{Epistemic Graph: A Plug-And-Play Module For Hybrid Representation Learning}

\author{Jin Yuan\\
Southeast University\\
Nanjing\\
\and
Yang Zhang \\
Lenovo Research \\
Beijing\\
\and
Yangzhou Du \\
Lenovo Research \\
Beijing\\
\and
Zhongchao Shi \\
Lenovo Research \\
Beijing\\
\and
Xin Geng\footnotemark[1] \\
Southeast University\\
Nanjing\\
\and
Jianping Fan \\
Lenovo Research \\
Beijing\\
\and
Yong Rui\footnotemark[1]\\
Lenovo Research \\
Beijing\\
}

\maketitle

\footnotetext[1]{Corresponding authors.}
\begin{abstract}
In recent years, deep models have achieved remarkable success in various vision tasks. However, their performance heavily relies on large training datasets. In contrast, humans exhibit hybrid learning, seamlessly integrating structured knowledge for cross-domain recognition or relying on a smaller amount of data samples for few-shot learning. Motivated by this human-like epistemic process, we aim to extend hybrid learning to computer vision tasks by integrating structured knowledge with data samples for more effective representation learning. Nevertheless, this extension faces significant challenges due to the substantial gap between structured knowledge and deep features learned from data samples, encompassing both dimensions and knowledge granularity. In this paper, a novel Epistemic Graph Layer (EGLayer) is introduced to enable hybrid learning, enhancing the exchange of information between deep features and a structured knowledge graph. Our EGLayer is composed of three major parts, including a local graph module, a query aggregation model, and a novel correlation alignment loss function to emulate human epistemic ability. Serving as a plug-and-play module that can replace the standard linear classifier, EGLayer significantly improves the performance of deep models. Extensive experiments demonstrates that EGLayer can greatly enhance representation learning for the tasks of cross-domain recognition and few-shot learning, and the visualization of knowledge graphs can aid in model interpretation.

\end{abstract}

\section{Introduction}
Over the past decade, deep models have achieved significant achievements in various vision tasks, relying on extensive data samples and complex model architectures \cite{bian2014knowledge, chen2014big, de2019deep, xie2021survey, hou2023learning}. In contrast, humans exhibit recognition ability with just a small number of samples, effortlessly achieving cross-domain recognition through a epistemic process known as hybrid learning. The core of hybrid learning lies in integrating structured knowledge with data samples to learn more effective representations (e.g., One can infer that the Chrysocyon brachyurus bears a striking visual resemblance to wolves and foxes, even if the observer has never encountered this species before). Motivated by this human capability, we sought to extend the principles of hybrid learning to deep learning methods. 

To represent the structured knowledge system of humans, a graph provides a direct and intuitive form of representation. In a graph, each node signifies a specific entity, and the relationships between these entities are encoded in the edge adjacency matrix. Compared with conventional knowledge fusion methods \cite{hu2016harnessing, allamanis2017learning, kodirov2017semantic, bansal2018zero, gurel2021knowledge, badreddine2022logic}, graph-based methods have two distinct advantages: 1. Node embeddings can encapsulate the general concept of an entity with rich knowledge; 2. Focusing on the relational adjacency matrix makes the graph representation inherently closer to human-structured knowledge.

One critical challenge in extending hybrid learning (e,g., incorporating knowledge graph into data-driven deep learning) is the mismatch between deep features (learned from data samples) and graph representations of structured knowledge. This mismatch can be categorized into two aspects: firstly, the deep features typically represent the visual distribution of a single image, while the structured knowledge graph contains the overall semantic knowledge which commonly share among substantial images, i.e., their information granularities are significantly different. Secondly, the deep features are usually in high dimensions, while the structured knowledge graph is a set of nodes and edges with much lower dimensions \cite{passban2021alp, rao2023parameter}. Existing methods mostly rely on a simple linear mapping \cite{lee2018multi, naeem2021learning} or matrix multiplication \cite{liang2018symbolic, chen2019multi, chen2020knowledge} to merge them, which could be ineffective and unstable.

For addressing the issue of information granularity mismatch, we intuitively propose local graph module that dynamically update a local prototypical graph by historical deep features. This module serves as a memory bank, enabling the transfer of deep features to the holistic visual graph. To fuse the input query samples with the local graph module, we devise a query aggregation model that incorporates the current deep feature to the local graph. We employ a Graph Neural Network (GNN) \cite{kipf2016semi, hamilton2017inductive, velivckovic2017graph} to aggregate information for both the local graph node and feature node, aligning them to the same dimension as the global graph. The final prediction is then based on the similarity between the local knowledge-enhanced deep features and the global node embeddings, mimicking the human process of using global knowledge to guide sample features. To strengthen the guidance process, a novel correlation alignment loss function is introduced to maintain linear consistency between the local graph and the global one by constraining the adjacency matrix from both cosine similarity and Euclidean space. Together, these three components constitute a well functional Epistemic Graph Layer (EGLayer).

The EGLayer stands out as a versatile plug-and-play hybrid learning module that seamlessly integrates into the majority of existing deep models, replacing the standard linear classifier. Our experiments on computer vision tasks, including cross-domain recognition and few-shot learning, have demonstrated the effectiveness of our proposed hybrid learning approach with EGLayer, showcasing substantial improvements in performance. Moreover, EGLayer has shown promising results compared to conventional knowledge integration methods. Additionally, the visualization of both local and global graphs provide valuable insights, contributing to model interpretation.

\section{Related Works}
Research on integrating human knowledge into deep models using graphs has garnered significant attention in recent years, primarily falling into two main streams: visual-guided graph representation learning and knowledge graph-guided visual feature learning.

\subsection{Visual-Guided Graph Representation Learning}
In this direction, works such as \cite{wang2018zero, chen2019multi, gao2019know, kampffmeyer2019rethinking, peng2019few, chen2020knowledge} often entail utilizing a fixed visual feature extractor and formulating a function to convert graph embeddings into visual features, subsequently integrating them. For instance, \cite{wang2018zero} constructs a Graph Convolutional Network (GCN) using the WordNet structure and trains it to predict visual classifiers pre-trained on ImageNet. By leveraging the relationships learned by GCN, it transfers knowledge to new class nodes, facilitating zero-shot learning. Building upon this, \cite{peng2019few} enhances the approach by introducing a knowledge transfer network, which replaces the inner product with cosine similarity of images. Additionally, \cite{chen2020knowledge} introduces a knowledge graph transfer network, which keeps the visual feature extractor fixed and employs three distance metrics to gauge the similarity of visual features.

\subsection{Knowledge Graph-Guided Visual Feature Learning}
Other works \cite{socher2013zero, norouzi2013zero, zhang2015zero, liang2018symbolic, monka2021learning, radford2021learning} commonly concentrate on knowledge graph-guided visual feature learning, favoring knowledge graphs for their perceived reliability over visual features. These approaches typically treat the knowledge graph either as a fixed external knowledge base or as high-level supervision for visual features. For instance, \cite{socher2013zero} employs a combination of dot-product similarity and hinge rank loss to learn a linear transformation function between the visual embedding space and the semantic embedding space, aiming to address issues in high-dimensional space. Notably, \cite{zhang2015zero} introduces a semantic similarity embedding method by representing target instances as a combination of a proportion of seen classes. They establish a semantic space where each novel class is expressed as a probability mixture of the projected source attribute vectors of the known classes. In a recent development, \cite{monka2021learning} leverages a knowledge graph to train a visual feature extractor using a contrastive knowledge graph embedding loss, showcasing superior performance compared to conventional methods.

To the best of our knowledge, existing works have made little effort to align the knowledge granularity between local image features and the global graph. Consequently, they often encounter challenges related to inefficient knowledge fusion and the underutilization of the knowledge embedded in the graph. This observation motivates us to explore a reliable and flexible knowledge graph projection method.

\begin{figure*}
	\centering
	\includegraphics[width=0.88\textwidth]{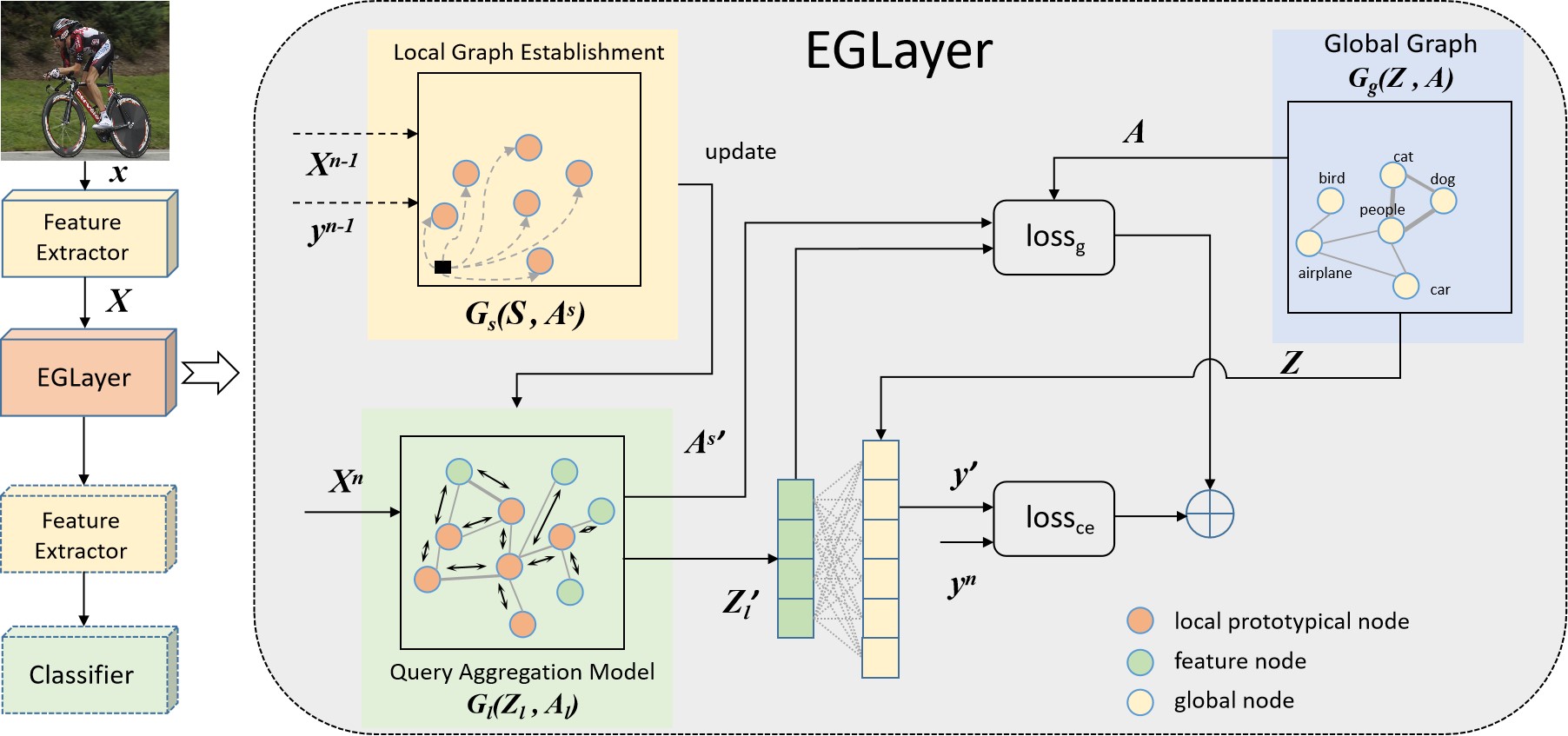}
	\caption{This figure illustrates the general framework of our proposed Epistemic Graph Layer. It can be inserted after any feature extractor layer to transfer the image feature dimension and granularity. In this paper, we primarily focus on replacing the standard linear classifier.}
	\label{fig1}
\end{figure*}

\section{Method}

For a typical classification task, we are provided with a dataset $ \mathcal{D} = (\boldsymbol{x}, \boldsymbol{y}) $ to train a model, where $ \boldsymbol{x} $ represents input images and $ \boldsymbol{y} $ denotes their respective labels. Initially, we employ a feature extractor $f_{\theta}$ to extract image features $\boldsymbol{X} \in \mathbb{R}^{D}$ from $\boldsymbol{x}$, with $\theta$ representing the learnable parameters. Subsequently, a classifier is employed to calculate the probability of each category based on the extracted features. Finally, the loss function (commonly used cross-entropy loss) between $ \boldsymbol{y}^{\prime} $ and $ \boldsymbol{y} $ are ultilized for optimization:

\begin{equation}
    \boldsymbol{X} = f_{\theta}(\boldsymbol{x}), \quad \boldsymbol{y}^{\prime} = \boldsymbol{W}\boldsymbol{X}, \quad \mathcal{L}_{sup} = loss_{ce}(\boldsymbol{y}, \boldsymbol{y}^{\prime}).
\end{equation}

Except for the labels of each instance, additional knowledge graphs could be available during model training. Assuming we have a global knowledge graph $\boldsymbol{G}_{g}$ (e.g., commonly obtained from manually annotated knowledge graphs or generated by entity embeddings from a large-scale corpus), the critical problem is how to integrate it to facilitate model training. We define $ \boldsymbol{G}_{g} = (\boldsymbol{Z}, \boldsymbol{A}) $, where $ \boldsymbol{Z} \in \mathbb{R}^{n \times d} $ represents the $ n $ nodes with $ d $-dimensional features, and $ \boldsymbol{A} \in \mathbb{R}^{n \times n} $ denotes the edges among the $ n $ nodes. 

\subsection{Linear Projection Layer}
\label{Sec31}
To integrate the knowledge graph $ \boldsymbol{G}_{g} $ into model training, the initial step is to project the visual features to the same dimension as the graph nodes, solving the previously discussed dimension mismatch problem. The most straightforward approach involves using a linear layer \cite{lee2018multi, naeem2021learning}, where $ \boldsymbol{W}_{p} \in \mathbb{R}^{d \times D}$ denotes the learnable mapping matrix. Subsequently, we can calculate the cosine similarity between $\boldsymbol{Z}'$ and the global graph node embedding $\boldsymbol{Z}_{i}$ to obtain the final prediction $ \boldsymbol{y}^{\prime} $, where $\langle\cdot, \cdot\rangle$ represents the cosine similarity of two vectors. The overall formulations are as follows:

\begin{equation}
    \boldsymbol{X} = f_{\theta}(\boldsymbol{x}), \quad \boldsymbol{Z^{\prime}} = \boldsymbol{W}_{p}\boldsymbol{X},   
\end{equation}

\begin{equation} 
    \quad \boldsymbol{y}^{\prime}=\frac{\exp \left(\left\langle \boldsymbol{Z}^{\prime}, \boldsymbol{Z}_{i} \right\rangle\right)}{\sum_{n} \exp \left(\left\langle \boldsymbol{Z}^{\prime}, \boldsymbol{Z}_{i}\right\rangle\right)}, \quad \mathcal{L}_{sup} = loss_{ce}(\boldsymbol{y}, \boldsymbol{y}^{\prime}). \label{eq5}
\end{equation}

\subsection{Epistemic Graph Layer}
To imitate the epistemic process observed in humans, we introduce a novel epistemic graph layer consisting of three key components. In this section, we provide a detailed introduction to these three modules. 

Firstly, the local graph module establishes a dynamically updated prototypical graph by historical features, serving as a memory bank that transfers instance-level features to a graph-level representation. Secondly, within the query aggregation model, the extracted features are injected into the obtained local graph to generate the query graph. This query graph is then input into a GNN to aggregate information for both feature and local graph nodes. This process ensures a natural dimension alignment between the local and global graphs, leading to the output of prediction logits. Finally, we propose an auxiliary correlation alignment loss by constraining the local and global correlation adjacency matrices. This constraint ensures linear consistency and comparable knowledge granularity between the local and global graphs, considering both cosine and Euclidean perspectives. The overall framework is shown in Figure~\ref{fig1}.

\subsubsection{Local Graph Establishment}
To address the challenge of knowledge granularity mismatch in deep learning when extending hybrid learning, we first construct a local graph $ \boldsymbol{G}_{l} = (\boldsymbol{Z}_{l}, \boldsymbol{A}_{l}) $ using the learned image features. Let $ \mathcal{D}_{k} $ denote the set of $ k $-th category samples. The local prototype $ \hat{\boldsymbol{S}} $ is initially obtained by averaging the features of each category:

\begin{equation}
\hat{\boldsymbol{S}}_{k} = \frac{1}{\left|\mathcal{D}_k\right|} \sum_{\left(\boldsymbol{x}_i, \boldsymbol{y}_i\right) \in \mathcal{D}_k} f_{\theta}\left(\boldsymbol{x}_i\right)
\end{equation}

To dynamically update the local prototype $ \boldsymbol{S}_{k} \in \mathbb{R}^{D} $, we employ exponential moving average scheme \cite{cai2021exponential, huang2021memory, xu2022graphical} in each iteration:

\begin{equation}
\boldsymbol{S}_{k}=\beta \boldsymbol{S}_{k}+(1-\beta)\hat{\boldsymbol{S}}_{k},
\end{equation}
where $ \beta $ is a hyperparameter controlling the balance between learning from recent features and preserving memories from early features. 

The local prototype $\boldsymbol{S}$ serves as the node embeddings of the local graph, acting as a local transfer station preserving historical visual features and aligning the granularity of the local graph with the semantic global graph. To enable interaction between the local graph and input query image features with batch size $ q $, we construct the updated local graph embedding $ \boldsymbol{Z}_{l} $:

\begin{equation}
\boldsymbol{Z}_{l}=[\underbrace{\boldsymbol{S}_{1} \boldsymbol{S}_{2} \cdots \boldsymbol{S}_{n}}_{\text {local prototypes}} \underbrace{\boldsymbol{X}_{1} \boldsymbol{X}_{2} \cdots \boldsymbol{X}_{q}}_{\text {query samples}} ]^{\mathrm{T}} . \label{eq6}
\end{equation}

\subsubsection{Query Aggregation Model}
To align the local graph with global graph in the same dimensional space for more effective utilization of global graph guidance, GNNs are employed through the aggregation operator. Prior to the aggregation process, it is imperative to define the adjacency matrix $ \boldsymbol{A}_{l} $. For each local prototype $ \boldsymbol{S} $ in the $ \boldsymbol{G}_{l} $, it is anticipated to aggregate information from closely related local graph nodes. We compute the adjacency matrix $ \boldsymbol{A}^{s} = \left(a_{i j}^{s}\right) \in \mathbb{R}^{n \times n} $ using the Gaussian kernel $ \mathcal{K}_G $ \cite{liu2019novel, wang2020learning, xu2022graphical}:

\begin{equation}
\begin{split}
    \boldsymbol{A}^{s} = \mathcal{K}_G\left(\boldsymbol{S}_i^{\mathrm{T}}, \boldsymbol{S}_j^{\mathrm{T}}\right) =\exp \left(-\frac{\left\|\boldsymbol{S}_i^{\mathrm{T}}-\boldsymbol{S}_j^{\mathrm{T}}\right\|_2^2}{2 \sigma^2}\right), \label{eq9}
\end{split}
\end{equation}
where $ \sigma $ is a hyperparameter controling the sparsity of $ \boldsymbol{A}^{s} $ that is set as 0.05 by default. Moreover, $ \boldsymbol{A}^{s} $ is a symmetric matrix ($ a^{s}_{i j} = a^{s}_{j i} $), allowing each node to both aggregate and transfer information. 

The query node $\boldsymbol{X}$ also needs to aggregate useful information from the prototypical nodes, and the aggregation matrix $\boldsymbol{A}^{xs} = \left(a^{xs}_{i j}\right) \in \mathbb{R}^{n \times q} $ is defined as:

\begin{equation}
\boldsymbol{A}^{xs} = \mathcal{K}_G\left(\boldsymbol{S}_i^{\mathrm{T}}, \boldsymbol{X}_j^{\mathrm{T}}\right)=\exp \left(-\frac{\left\|\boldsymbol{S}_i^{\mathrm{T}}-\boldsymbol{X}_j^{\mathrm{T}}\right\|_2^2}{2 \sigma^2}\right).
\end{equation}

Subsequently, the adjacency matrix $ \boldsymbol{A}_{l} $ is calculated as follows:

\begin{equation}
\boldsymbol{A}_{l}=\left[\begin{array}{cc}
\boldsymbol{A}^{s} & \boldsymbol{A}^{xs} \\
\boldsymbol{A}^{xs\mathrm{T}} & \boldsymbol{E}
\end{array}\right], \label{eq11}
\end{equation}
where $ \boldsymbol{E} $ represents the identity matrix since query features are not allowed to interact with each other.

With the local graph embedding $ \boldsymbol{Z}_{l} $ and adjacency matrix $ \boldsymbol{A}_{l} $, we exploit GCN \cite{estrach2014spectral, kipf2016semi} to perform the aggregation operation:

\begin{equation}
	\boldsymbol{H}^{(m+1)}=\sigma\left(\boldsymbol{\tilde{D}}_{l}^{-\frac{1}{2}} \boldsymbol{\tilde{A}}_{l} \boldsymbol{\tilde{D}}_{l}^{-\frac{1}{2}} \boldsymbol{H}^{(m)} \boldsymbol{W}^{(m)}\right),
\end{equation}
where $ \boldsymbol{\tilde{A}}_{l} $ is the local correlation matrix $ \boldsymbol{A}_{l} $ with self-connections, and $ \boldsymbol{\tilde{D}}_{l} $ is the degree matrix of $ \boldsymbol{\tilde{A}}_{l} $. $ \boldsymbol{W}^{(m)} $ denotes the learnable matrix in $ m $-th layer, while $ \sigma $ is the activation function. Here, we take the local graph embedding $ \boldsymbol{Z}_{l} $ as the first layer input of $\boldsymbol{H}^{(m)}$, and the final aggregated node representation $ \boldsymbol{H}^{(m+1)} $ are defined as $ \boldsymbol{Z}_{l}^{\prime} $, which consists of $ \boldsymbol{S}^{\prime} $ and $\boldsymbol{X}^{\prime} $ as Eq.~\ref{eq6}.

Finally, we exploit Eq.~\ref{eq5} to calculate the output predictions by $\boldsymbol{X}^{\prime} $ and global node embedding $ \boldsymbol{Z} $. 

\subsubsection{Correlation Alignment Loss}
To ensure sufficient and consistent guidance from the global graph, we deliberately impose constraints on the local adjacency matrix. Nevertheless, the local adjacency matrix is fixed in each training iteration, as $ \boldsymbol{A}^{s} $ is solely dependent to the local graph embedding $ \boldsymbol{S} $, which is updated in advance of each iteration. Consequently, we introduce an extra learnable matrix $\boldsymbol{W}_{a}$ for $ \boldsymbol{A}^{s} $ to obtain the amended adjacency matrix:

\begin{equation}
\boldsymbol{A}^{s\prime} = \left(a_{i j}^{s\prime}\right) \in \mathbb{R}^{n \times n} = \boldsymbol{W}_{a}\boldsymbol{A}^{s}_{i, j}=\boldsymbol{W}_{a}\mathcal{K}_G\left(\boldsymbol{S}_i^{\mathrm{T}}, \boldsymbol{S}_j^{\mathrm{T}}\right) .
\end{equation}

Then, the adjacency matrix in Eq.~\ref{eq11} is finalized as:

\begin{equation}
\boldsymbol{A}_{l}=\left[\begin{array}{cc}
\boldsymbol{A}^{s\prime} & \boldsymbol{A}^{xs} \\
\boldsymbol{A}^{xs\mathrm{T}} & \boldsymbol{E}
\end{array}\right].
\end{equation}

Accordingly, we build an auxiliary loss function by optimize $ \boldsymbol{A}^{s\prime} $ to the global adjacency matrix $ \boldsymbol{A} $:

\begin{equation}
\begin{split}
\mathcal{L}_{a}(\boldsymbol{A}, \boldsymbol{A}^{s\prime}) = - \frac{1}{n^{2}} \sum_{i=1}^{n} \sum_{j=1}^{n} [a_{i j} \log \left(\sigma\left(a^{s\prime}_{i j}\right)\right)\\
+\left(1-a_{i j}\right) \log \left(1-\sigma\left(a^{s\prime}_{i j}\right)\right)] , \label{eq15}
\end{split}
\end{equation}
where $\sigma(\cdot)$ is sigmoid function and $ \mathcal{L}_{a} $ could be viewed as a binary cross-entropy loss for each correlation value with soft labels. 

Moreover, since $ \boldsymbol{A} $ and $ \boldsymbol{A}^{s\prime} $ both come from Euclidean space, we design a new regularization term based on cosine similarity to make learned embedding $ \boldsymbol{S}^{\prime} $ more distinctive. The regularization is calculated as: 

\begin{equation}
\begin{split}
\mathcal{L}_{reg}(\boldsymbol{S}^{\prime}) &= \| \left\langle \boldsymbol{S}^{\prime}, \boldsymbol{S}^{\prime\mathrm{T}} \right\rangle \|_{2}  = \|\boldsymbol{C}\|_{2} \\
&= \|\left(c_{i j}\right) \in \mathbb{R}^{n \times n}\|_{2} = \sqrt{\sum_{i=1}^n \sum_{j=1}^n c_{i j}^2}
\end{split}
\end{equation}

Finally, the overall loss function combines supervised loss and correlation alignment loss:

\begin{equation}
\mathcal{L} =  \mathcal{L}_{sup} + \alpha \mathcal{L}_{g} = \mathcal{L}_{sup} + \alpha_{1} \mathcal{L}_{a} + \alpha_{2} \mathcal{L}_{reg}. \label{eq15}
\end{equation}

\begin{table*}[h] 
	\caption{Comparison experiments on Office-31 dataset}
	\label{Table1}
	\centering
	\begin{tabular}{cccccccc}
	\toprule
	Methods    &  A→W & D→W & W→D & A→D & D→A & W→A & Average \\
	\midrule
    ResNet50  & 65.41 & 79.25 & 91.00 & 70.00 & 44.68 & 50.38 & 66.79  \\
    ResNet50 + LPLayer & 67.92 & 85.53 & 94.00 & 71.00 & 53.62 & 56.22 & 71.38  \\
    ResNet50 + EGLayer & \bf{70.44} & \bf{90.57} & \bf{96.00} & \bf{77.00} & \bf{56.96} & \bf{57.87} & \bf{74.81} \\
    \bottomrule
	\end{tabular}
\end{table*}

\begin{table*} 
	\caption{Comparison experiments on Office-Home dataset}
	\label{Table2}
        \setlength\tabcolsep{3pt} 
	\centering
	\begin{tabular}{cccccccccccccc}
	\toprule
	Methods    &  A→C & A→P & A→R & C→A & C→P & C→R & P→A & P→C & P→R & R→A & R→C & R→P & Average \\
	\midrule
    ResNet50  & 40.42 & \bf{59.48} & \bf{69.10} & 45.07 & \bf{56.55} & \bf{60.13} & 39.71 & 39.86 & 68.09 & 58.64 & 43.60 & 73.64 & 54.52 \\
    ResNet50 + LPLayer & 40.78 & 28.69 & 66.43 & 40.48 & 32.88 & 43.04 & 56.68 & \bf{44.81} & \bf{69.03} & 65.08 & \bf{49.79} & 52.58 & 49.19 \\
   ResNet50 + EGLayer & \bf{41.81} & 57.95 & 65.74 & \bf{53.36} & 53.35 & 56.34 & \bf{62.52} & 41.67 & 68.28 & \bf{70.33} & 45.54 & \bf{73.67} & \bf{57.55} \\
    \bottomrule
	\end{tabular}
\end{table*}

\section{Experiments}

As previously discussed, our proposed EGLayer serves as a plug-and-play module capable of enhancing various types of deep models by seamlessly replacing their standard linear classifiers. To assess the effectiveness of our knowledge guidance and extrapolation, we conduct extensive evaluations on several challenging tasks, including cross-domain classification, open-set domain adaptation, and few-shot learning.

The establishment of the global knowledge graph encompasses various available schemes. The co-occurrence graph \cite{duque2018co, chen2019multi, wang2020multi} represents the frequency of two classes occurring together but is not well-suited for single-label tasks and heavily relies on the dataset size. Another option is the pre-defined knowledge graph \cite{lin2015learning, toutanova2016compositional, krishna2017visual}, constructed using manually labeled relational datasets or knowledge bases. In our approach, we opt for a simpler solution by employing word embeddings from GloVe \cite{pennington2014glove} and Eq.~\ref{eq9} to derive node embeddings and adjacency matrices. This adaptive approach does not require additional sources of knowledge and easy to utilize.

Notably, in our experiments, we solely leverage class information from the training set, refraining from integrating any novel classes information into the global knowledge graph. In the context of open-set domain adaptation, our approach begins by training the model on the source domain, emphasizing source classes. Subsequently, we apply a threshold to filter out images not belonging to known classes within the source domain, categorizing them as outlier classes. In the realm of few-shot learning, our method trains the feature extractor and constructs both global and local graphs based on the base classes. During validation and testing phases, the trained feature extractor is employed to extract image features for the few-shot images associated with the novel class. Following this, unlabeled test images are compared to these few-shot features using cosine similarity to determine their respective classes.

\subsection{Cross-Domain Classification}
\subsubsection{Datasets}
In this experiment, we train the model on the source domain and then perform classification directly on the target domain without utilizing any target domain data. We conduct experiments on two datasets, namely Office-31 \cite{saenko2010adapting} and Office-Home \cite{venkateswara2017deep}. The Office-31 dataset comprises of 4,652 images from 31 categories and is partitioned into three domains: \emph{Amazon} (A), \emph{Dslr} (D), and \emph{Webcam} (W). The Office-Home dataset has 15,500 images with 65 categories and is divided into four domains: \emph{Art} (A), \emph{Clipart} (C), \emph{Product} (P), and \emph{Real World} (R).

\begin{figure*}
	\centering
	\includegraphics[width=\textwidth]{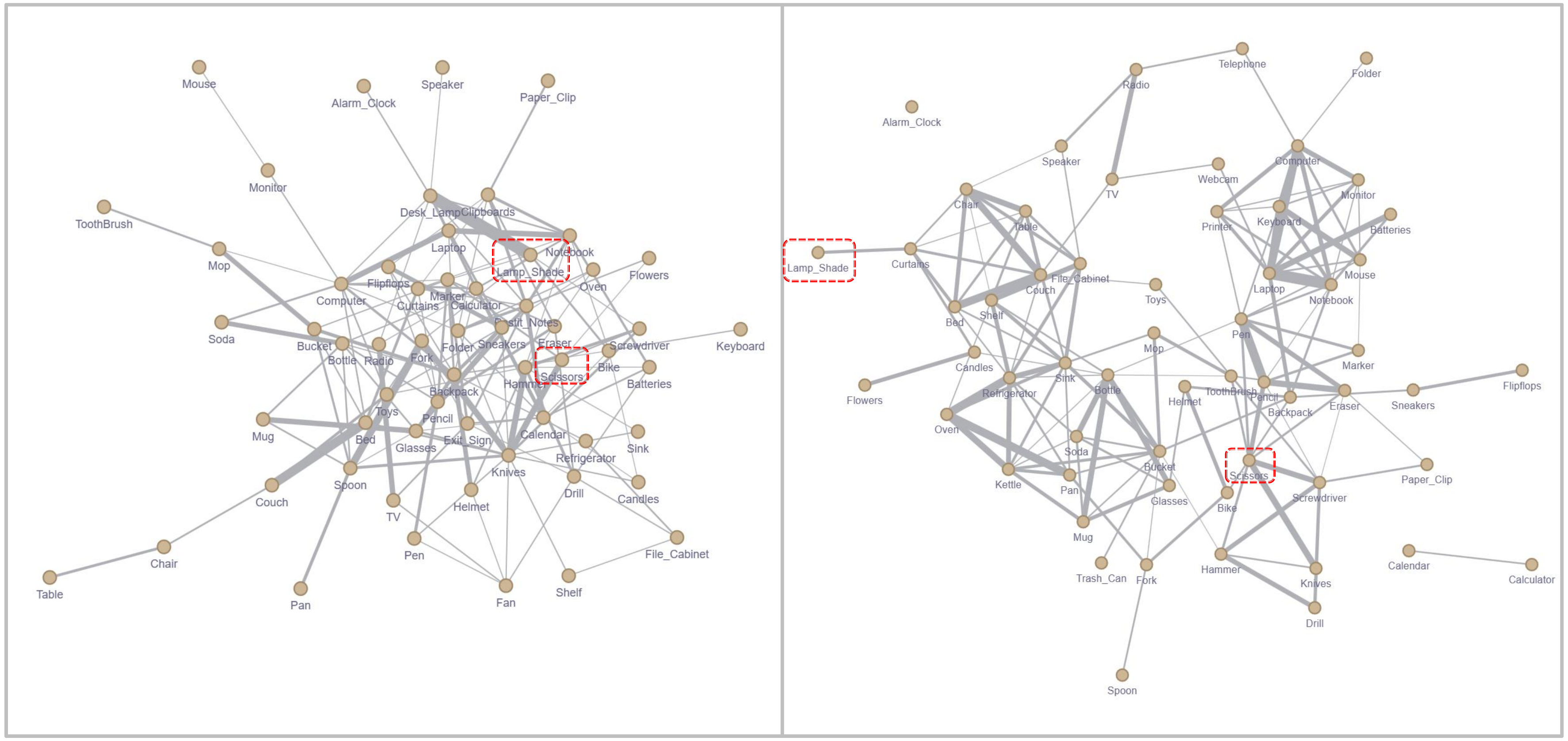}
	\caption{The left visualized graph is enhanced local graph, and the right is global graph. These experiments are conducted in Office-Home datasets of 65 classes in \emph{Clipart} domain. We have highlighted two typical nodes: the \emph{Lamp Shade} is visually similar to \emph{Desk Lamp} while the \emph{Scissors} is semantically closer to stationery objects.}
	\label{fig2}
\end{figure*}

\subsubsection{Comparison Results}
Table \ref{Table1} and Table \ref{Table2} showcase the results of our experiments with various model settings. ResNet50 \cite{he2016deep} denotes ResNet50 backbone paired with a standard linear classifier. ResNet50 + LPLayer signifies the ResNet50 backbone with the linear projection layer described in Section~\ref{Sec31}. ResNet50 + EGLayer is the ResNet50 backbone equipped with our proposed epistemic graph layer. The sole distinction among the three models lies in the classifier, enabling a fair and direct comparison. 

On average, ResNet50 + LPLayer outperforms ResNet50 by 4.59\% on Office-31. Furthermore, ResNet50 + EGLayer exhibits an additional performance gain of 3.43\%, securing the best results across all cases. Surprisingly, ResNet50 + LPLayer shows an obvious performance drop on Office-Home by 5.33\%, possibly due to insufficient knowledge integration. Conversely, ResNet50 + EGLayer achieves a noteworthy improvement by 3.03\%. Notably, the largest margin is reported in the D→W task on Office-31, where ResNet50 + EGLayer elevates the results from 79.25\% to 90.57\%, marking an impressive increase of 11.32\%. These findings underscore the EGLayer's capacity to learn a superior representation.

\subsubsection{Visualization of Graphs}
We present visualizations of two graphs, namely the enhanced local graph and the global graph. For clarity, we display only the top 150 edges with strong relationships, where the thickness of each edge corresponds to a higher relational edge value. (See Appendix for more details.)

The enhanced local graph primarily encompasses knowledge derived from visual sources, while the global graph incorporates a broader spectrum of semantic knowledge. Illustrated in Figure~\ref{fig2}, we emphasize two characteristic nodes. The \emph{Scissors} node in the global graph is proximate to two conceptual categories, namely tools and stationeries. The tools category includes \emph{Knives}, \emph{Hammer}, and \emph{Screwdriver}, while the stationeries category comprises \emph{Eraser}, \emph{Pencil}, and \emph{Pen}. In the enhanced local graph, \emph{Scissors} is exclusively associated with the typical tools category, owing to their shared metallic appearance.

Another noteworthy node is \emph{Lamp Shade}, which exhibits a high association with \emph{Desk Lamp} due to the frequent pairing of \emph{Lamp Shade} images with lamps. Interestingly, these two nodes lack an edge in the global graph, a phenomenon that could be attributed to the semantic emphasis on \emph{Lamp Shade} as a shade rather than a lamp.

\subsection{Open-Set Domain Adaptation}

\subsubsection{Implementation Details}

\begin{table*}[h] 
	\caption{Universal domain adaptation experiments on Office-31 dataset}
	\label{Table3}
         \setlength\tabcolsep{3pt} 
	\centering
	\begin{tabular}{cccccccc}
	\toprule
	Methods    &  A→W & D→W & W→D & A→D & D→A & W→A & Average \\
	\midrule
 DANN \cite{ganin2016domain} & 80.65 & 80.94 & 88.07 & 82.67 & 74.82 & 83.54 & 81.78 \\
 RTN \cite{long2016unsupervised} & \bf{85.70} & 87.80 & 88.91 & 82.69 & 74.64 & 83.26 & 84.18 \\
 IWAN \cite{zhang2018importance} & 85.25 & 90.09 & 90.00 & 84.27 & 84.22 & 86.25 & 86.68 \\
 PADA \cite{zhang2018importance} & 85.37 & 79.26 & 90.91 & 81.68 & 55.32 & 82.61 & 79.19 \\
 ATI \cite{panareda2017open} & 79.38 & 92.60 & 90.08 & 84.40 & 78.85 & 81.57 & 84.48 \\
 OSBP \cite{saito2018open} & 66.13 & 73.57 & 85.62 & 72.92 & 47.35 & 60.48 & 67.68 \\
UAN \cite{you2019universal} & 77.16 & \bf{94.54} & \bf{95.48} & 78.71 & 84.47 & 82.14 & 85.42 \\
\midrule
UAN + LPLayer & 83.69 & 91.20 & 95.17 & 84.90 & 84.93 & 84.24 & 87.36 \\
UAN + EGLayer & 83.51 & 94.23 & 94.34 & \bf{86.11}  & \bf{87.88} & \bf{88.26} & \bf{89.06} \\
    \bottomrule
	\end{tabular}
\end{table*}

\begin{table*}
	\caption{Universal domain adaptation experiments on Office-Home dataset}
	\label{Table4}
        \setlength\tabcolsep{2.5pt} 
	\centering
	\begin{tabular}{cccccccccccccc}
	\toprule
	Methods    &  A→C & A→P & A→R & C→A & C→P & C→R & P→A & P→C & P→R & R→A & R→C & R→P & Average \\
	\midrule
 DANN \cite{ganin2016domain} & 56.17 & 81.72 & 86.87 & 68.67 & 73.38 & 83.76 & 69.92 & 56.84 & 85.80 & 79.41 & 57.26 & 78.26 & 73.17 \\
 RTN \cite{long2016unsupervised} & 50.46 & 77.80 & 86.90 & 65.12 & 73.40 & 85.07 & 67.86 & 45.23 & 85.50 & 79.20 & 55.55 & 78.79 & 70.91 \\
 IWAN \cite{zhang2018importance} & 52.55 & 81.40 & 86.51 & 70.58 & 70.99 & 85.29 & 74.88 & 57.33 & 85.07 & 77.48 & 59.65 & 78.91 & 73.39 \\
 PADA \cite{zhang2018importance} & 39.58 & 69.37 & 76.26 & 62.57 & 67.39 & 77.47 & 48.39 & 35.79 & 79.60 & 75.94 & 44.50 & 78.10 & 62.91 \\
ATI \cite{panareda2017open} & 52.90 & 80.37 & 85.91 & 71.08 & 72.41 & 84.39 & 74.28 & 57.84 & 85.61 & 76.06 & 60.17 & 78.42 & 73.29 \\
OSBP \cite{saito2018open} & 47.75 & 60.90 & 76.78 & 59.23 & 61.58 & 74.33 & 61.67 & 44.50 & 79.31 & 70.59 & 54.95 & 75.18 & 63.90 \\
UAN \cite{you2019universal} & 65.92 & 79.82 & 88.09 & 71.99 & 75.11 & 84.54 & 77.56 & \bf{64.16} & 89.06 & \bf{81.92} & 65.87 & 83.80 & 77.32 \\ 

\midrule
UAN + LPLayer & \bf{67.43} & 81.64 & 88.97 & 76.19 & 81.58 & 87.29 & 79.86 & 63.11 & 88.73 & 79.70 & \bf{68.62} & 84.07 & 78.93 \\
  UAN + EGLayer & 66.47 & \bf{84.53} & \bf{92.36} & \bf{80.97} & \bf{82.79} & \bf{89.40} & \bf{80.12} & 63.35 & \bf{91.98} & 79.48 & 64.54 & \bf{85.43} & \bf{80.12} \\
    \bottomrule
	\end{tabular}
\end{table*}

In this subsection, we conduct experiments on open-set domain adaptation tasks, where the source and target domains have some shared and some private categories. We adopt the task definition proposed in \cite{you2019universal}. Specifically, we denote the label sets of the source and target domains as $\mathcal{C}_s$ and $\mathcal{C}_t$, respectively, and $\mathcal{C}=\mathcal{C}_s \cap \mathcal{C}_t$ represents the set of shared categories. Furthermore, $\overline{\mathcal{C}}_s=\mathcal{C}_s \backslash \mathcal{C}$ and $\overline{\mathcal{C}}_t=\mathcal{C}_t \backslash \mathcal{C}$ represent the private categories in the source and target domains, respectively. We can then quantify the commonality between the two domains as:
\begin{equation}
    \xi=\frac{\left|\mathcal{C}_s \cap \mathcal{C}_t\right|}{\left|\mathcal{C}_s \cup \mathcal{C}_t\right|}. 
\end{equation}

For the Office-31, we choose 10 categories as shared categories $\mathcal{C}$, the following 10 categories as source private categories $\mathcal{C}_s$, and the remaining categories as target private categories $\mathcal{C}_t$. For the Office-Home, we take the first 10 categories as $\mathcal{C}$, the next 5 categories $\mathcal{C}_s$, and the rest as $\mathcal{C}_t$. As a result, we obtain $\xi$ values of 0.32 and 0.15 for the Office-31 and Office-Home, respectively. (See Appendix for more experiments.)

\subsubsection{Comparison Results}
We summarize the results in Table \ref{Table3} and Table \ref{Table4}. To comprehensively assess the effect of knowledge integration, we replace the linear classifier in UAN \cite{you2019universal} with LPLayer and EGLayer, resulting in UAN+LPLayer and UAN+EGLayer, respectively. 

In the open-world setting, the integration of knowledge emerges as a pivotal factor for performance enhancement. On average, UAN + LPLayer demonstrates 1.94\% and 1.61\% improvements over baseline UAN on Office-31 and Office-Home datasets. The proposed UAN + EGLayer further elevates the results by 1.70\% and 1.19\% in comparison to UAN + LPLayer, indicating that EGLayer exhibits superior generalization capabilities in contrast to conventional linear knowledge fusion methods. Notably, both knowledge-based approaches show more pronounced improvements in challenging tasks (i.e. tasks with low accuracy), such as D→A and A→D. In general, UAN + EGLayer outperforms all competitors and achieves state-of-the-art performance in the open-world setting.

\begin{figure}
	\centering
	\includegraphics[width=\columnwidth]{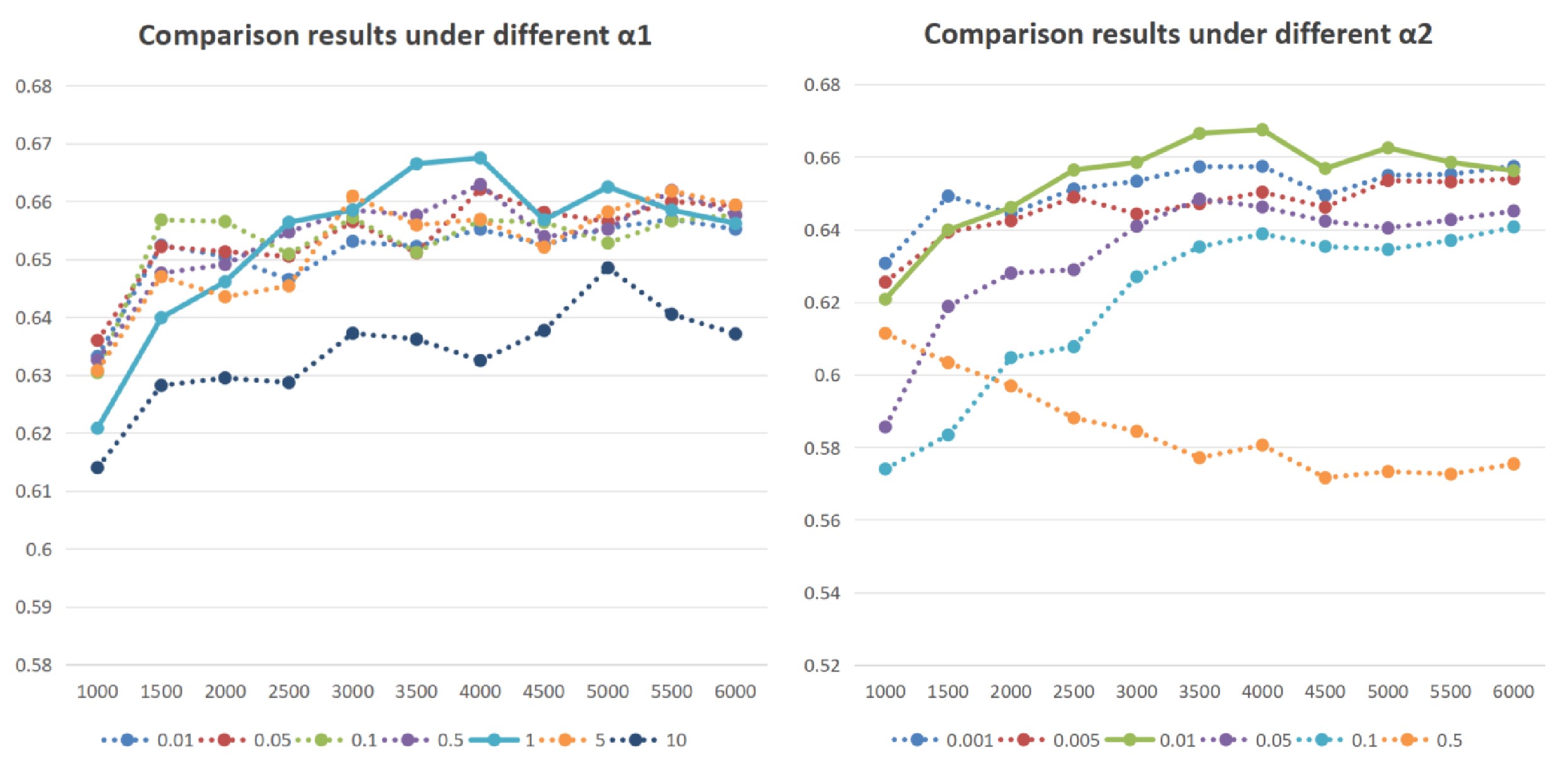}
	\caption{The results of the comparison are presented under varying values of $\alpha_{1}$ and $ \alpha_{2} $. The validation results demonstrate that $\alpha_{1}=1.0$ and $\alpha_{2}=0.01$ yields the best performance as compared to higher and lower values of $\alpha_{1}$ and $\alpha_{2}$. }
	\label{fig4}
\end{figure}

\subsubsection{Correlation Loss Study}
We conduct experiments to determine the optimal values for $\alpha_{1}$ and $\alpha_{2}$ in Eq.~\ref{eq15}. The results on the validation set from \emph{art} to \emph{clipart}, with 6,000 iterations, are depicted in Figure~\ref{fig4}. In the left chart, we fix $\alpha_{2}$ and train the model with $\alpha_{1}$ values of 0.01, 0.05, 0.1, 0.5, 1, 5, and 10. In the right chart, we maintain $\alpha_{1}$ constant and train the model with $\alpha_{2}$ values of 0.001, 0.005, 0.01, 0.05, 0.1, and 0.5.

We observe that all experimental settings reach their peak performance between 3,000 and 4,500 iterations. For $\alpha_{1}$, excessively large values ($\alpha_{1}=10$) result in significantly poorer performance. The performance of the other weights are similar, peaking at around 66\%. Notably, the validation set result with a weight of 1.0 significantly outperforms other settings, leading us to choose $\alpha_{1}=1.0$.

Regarding $\alpha_{2}$, we notice that even $\alpha_{2}=0.5$ lead to performance degradation, indicating that excessive regularization could impede the model's learning ability. The experimental performance of the other settings are relatively close, with only the 0.01 version exceeding 66\%. Consequently, we select $\alpha_{2}$ as 0.01.

\subsection{Few-Shot Learning}
\subsubsection{Datasets}
We evaluate the few-shot learning task on two datasets. The miniImageNet \cite{vinyals2016matching} is sampled from ImageNet \cite{russakovsky2015imagenet} of 100 classes. 64 classes are used for training, the rest 16 and 20 classes are used for validation and testing, respectively. Each class contains 600 images resized to 84 $ \times $ 84 resolution. The tieredImageNet \cite{ren2018meta} is a larger datasets consisting of 608 classes sampled from ImageNet \cite{russakovsky2015imagenet} too. All classes are divided 351, 97, 160 classes for training, validation and testing. Different from miniImageNet, tieredImageNet is more challenging owing to the long semantic distance between base and novel classes. (See Appendix for more implementation details and experiments.)

\begin{table}[h] \scriptsize
	\caption{Comparison with state-of-the-art methods on miniImageNet dataset.}
	\label{Table5}
	\centering
	\begin{tabular}{cccc}
	\toprule
	Methods    & Backbone & 1-shot & 5-shot \\
	\midrule
	SNAIL \cite{mishra2017simple} & ResNet-12 & 55.71 ± 0.99 & 68.88 ± 0.92 \\
	AdaResNet \cite{munkhdalai2018rapid} & ResNet-12 & 56.88 ± 0.62 & 71.94 ± 0.57 \\
	TADAM \cite{oreshkin2018tadam} & ResNet-12 & 58.50 ± 0.30 & 76.70 ± 0.30 \\
	MTL \cite{sun2019meta} & ResNet-12 & 61.20 ± 1.80 & 75.50 ± 0.80 \\
	 MetaOptNet \cite{lee2019meta} & ResNet-12 & 62.64 ± 0.61 & 78.63 ± 0.46 \\
	ProtoNets + TRAML \cite{li2020boosting} & ResNet-12 & 60.31 ± 0.48 & 77.94 ± 0.57 \\
 BOIL \cite{oh2021boil} & ResNet-12 & - & 71.30 ± 0.28 \\
 DAM \cite{zhou2022meta} & ResNet-12 & 60.39 ± 0.21 & 73.84 ± 0.16 \\
    \midrule
          Matching Networks \cite{vinyals2016matching} & ConvNet-4 & 45.73 ± 0.19 & 57.80 ± 0.18 \\
          Matching Networks + LPLayer & ConvNet-4 & 47.87 ± 0.19 & 57.84 ± 0.18 \\
          Matching Networks + EGLayer & ConvNet-4 & \bf{50.48 ± 0.20} & \bf{61.29 ± 0.17} \\
           \midrule
            Prototypical Networks \cite{snell2017prototypical} & ConvNet-4 & 49.45 ± 0.20 & 66.38 ± 0.17 \\
          Prototypical Networks + LPLayer & ConvNet-4 & 49.67 ± 0.20 & 66.66 ± 0.17 \\
          Prototypical Networks + EGLayer & ConvNet-4 & \bf{50.30 ± 0.20} & \bf{67.88 ± 0.16} \\
                   \midrule
    Classifier-Baseline \cite{chen2021meta} & ResNet-12 & 58.91 ± 0.23 & 77.76 ± 0.17 \\
    Classifier-Baseline + LPLayer & ResNet-12 & 60.96 ± 0.23 & 78.07 ± 0.17 \\
    Classifier-Baseline + EGLayer & ResNet-12 & \bf{61.53 ± 0.27} & \bf{78.84 ± 0.21} \\
               \midrule
    Meta-Baseline \cite{chen2021meta} & ResNet-12 & 63.17 ± 0.23 & 79.26 ± 0.17 \\
    Meta-Baseline + LPLayer & ResNet-12 & 62.27 ± 0.23 & 77.63 ± 0.17 \\
    Meta-Baseline + EGLayer & ResNet-12 & \bf{63.55 ± 0.26} & \bf{79.78 ± 0.54} \\
    \bottomrule
	\end{tabular}
\end{table}

\begin{table}[h] \scriptsize
	\caption{Comparison with state-of-the-art methods on tieredImageNet dataset}
	\label{Table6}
	\centering
	\begin{tabular}{cccc}
	\toprule
	Methods    & Backbone & 1-shot & 5-shot \\
	\midrule
        MAML \cite{finn2017model} & ConvNet-4 & 51.67 ± 1.81 & 70.30 ± 1.75 \\
        Relation Networks \cite{sung2018learning} & ConvNet-4 & 54.48 ± 0.93 & 71.32 ± 0.78 \\
	MetaOptNet \cite{lee2019meta} & ResNet-12 & 65.99 ± 0.72 & 81.56 ± 0.53 \\
  BOIL \cite{oh2021boil} & ResNet-12 & 48.58 ± 0.27 & 69.37 ± 0.12 \\
 DAM \cite{zhou2022meta} & ResNet-12 & 64.09 ± 0.23 & 78.39 ± 0.18 \\
A-MET \cite{zheng2023detach} & ResNet-12 & 69.39 ± 0.57 & 81.11 ± 0.39 \\
    \midrule
         Matching Networks \cite{vinyals2016matching} & ConvNet-4 & 41.99 ± 0.19 &  52.70 ± 0.19 \\
          Matching Networks + LPLayer & ConvNet-4 & 42.61 ± 0.20 & 52.91 ± 0.19 \\
          Matching Networks + EGLayer & ConvNet-4 & \bf{45.87 ± 0.22} & \bf{59.90 ± 0.19} \\
           \midrule
           Prototypical Networks \cite{snell2017prototypical} & ConvNet-4 & 48.65 ± 0.21 & 65.55 ± 0.19 \\
          Prototypical Networks + LPLayer &  ConvNet-4 & 48.97 ± 0.21 & 65.52 ± 0.19 \\
          Prototypical Networks + EGLayer &  ConvNet-4 & \bf{50.17 ± 0.22} & \bf{68.42 ± 0.18} \\
          \midrule
    Classifier-Baseline \cite{chen2021meta} & ResNet-12 & 68.07 ± 0.26 & 83.74 ± 0.18 \\
    Classifier-Baseline + LPLayer & ResNet-12 & 68.28 ± 0.26 & 83.04 ± 0.18 \\
    Classifier-Baseline + EGLayer & ResNet-12 & \bf{69.38 ± 0.53} & \bf{84.38 ± 0.59} \\
    \midrule
    Meta-Baseline \cite{chen2021meta} & ResNet-12 & 68.62 ± 0.27 & 83.74 ± 0.18 \\
    Meta-Baseline + LPLayer & ResNet-12 & 69.16 ± 0.56 & 82.64 ± 0.41 \\
    Meta-Baseline + EGLayer & ResNet-12 & \bf{69.74 ± 0.56} & \bf{83.94 ± 0.58} \\
    \bottomrule
	\end{tabular}
\end{table}

\subsubsection{Comparison Results}
We conduct a comparative analysis of our proposed method against mainstream approaches, and the results are presented in Tables \ref{Table5} and \ref{Table6}. All reported results represent the average 5-way accuracy with a 95\% confidence interval. To validate the lightweight and plug-and-play nature of our method, we implement our methods with four prevailing baselines Matching Networks \cite{vinyals2016matching}, Prototypical Networks \cite{snell2017prototypical}, Classifier-Baseline \cite{chen2021meta}, and Meta-Baseline \cite{chen2021meta}.

For miniImageNet, LPLayer versions exhibit marginal improvements over the baseline, and inserting a LPLayer even causes a slight performance decline in Meta-Baseline. In contrast, EGLayer consistently achieves stable improvements across all results. Especially for Matching Networks and Classifier-Baseline, EGLayer gains 4.75\%$/$3.49\% and 2.62\%$/$1.08\% promotion. 

For tieredImageNet, compared with LPLayer, EGLayer enables a more effective and reliable knowledge injection, resulting in significant advantages in different settings. In detail, the EGLayer demonstrates improvements of 3.88\%$/$7.20\%, 1.52\%$/$2.87\% with Matching Networks and Prototypical Networks, respectively. For Classifier-Baseline and Meta-Baseline, EGLayer also exhibits a remarkable advantages in 1-shot setting with 1.31\% and 1.12\% performance enhancements. 

\section{Conclusions}

This paper introduces a novel EGLayer to enable hybrid learning, enhancing the effectiveness of information exchange between local deep features and a structured global knowledge graph. EGLayer serves as a plug-and-play module, seamlessly replacing the standard linear classifier. Its integration significantly improves the performance of deep models by effectively blending structured knowledge with data samples for deep learning. Our extensive experiments demonstrate that the proposed hybrid learning approach EGLayer substantially enhances representation learning for cross-domain recognition and few-shot learning tasks. Additionally, the visualization of knowledge graphs proves to be an effective tool for model interpretation.

{\small

}

\appendix

\section{Visualization}
In this section, we present three graphs of the Office-31 and Office-Home datasets: the local visual graph, enhanced local graph, and global graph. Figure~\ref{fig1} displays these graphs, with the upper line representing the results of the Office-Home dataset and the bottom line representing the Office-31 dataset. The first column displays the local visual graph, the second column displays the enhanced local graph, and the right column displays the global graph. In these graphs, thicker edges indicate stronger relations, while node size remains constant. To maintain clarity and prevent clutter resulting from an excess of edges, we show the top-150 edges in the Office-Home dataset and top-70 edges in the Office-31 dataset. Additionally, we have highlighted two nodes in each graph to demonstrate the differences among the three graphs.

In the Office-Home dataset, the \emph{Scissors} node in the global graph is positioned near two types of concepts: tools and stationery. The typical tools include \emph{Knives}, \emph{Hammer}, and \emph{Screwdriver}, while the stationery items encompass \emph{Eraser}, \emph{Pencil}, and \emph{Pen}. In the local visual graph, \emph{Scissors} is primarily associated with the typical tools category due to their similar metallic appearance. In the enhanced local graph, \emph{Scissors} has features from both the semantic and visual graphs. Specifically, it establishes robust connections with \emph{Knives} and \emph{Screwdriver}, and a comparatively thinner edge with \emph{Pen}. Another noteworthy node is \emph{Mop}, which is visually linked to \emph{Toothbrush}, \emph{Bucket}, \emph{Curtains}, and other objects in the visual graph. However, in the semantic graph, it is only related to \emph{Toothbrush}, \emph{Bucket}, and \emph{Sink}. As a result, in the enhanced visual graph, \emph{Mop} is positioned closer to the semantic graph with three edges connecting it to \emph{Toothbrush}, \emph{Bucket}, and \emph{Bottle}.

In the Office-31 dataset, the \emph{mouse} node lacks connections to other nodes in the local visual graph due to its distinctive appearance. However, it has several neighbors, including \emph{keyboard}, \emph{laptop computer}, and others in global graph. In the enhanced local graph, \emph{mouse} begins to establish some edges with other nodes, confirming the guidance of the global graph. For \emph{ruler}, it maintains an edge with \emph{pen} in the local visual graph, while it does not have an edge in the global graph. In the enhanced local graph, it still retains an edge with \emph{pen}, showcasing the preservation of visual information within the enhanced local graph.

\begin{figure*}
	\centering
	\includegraphics[width=\textwidth]{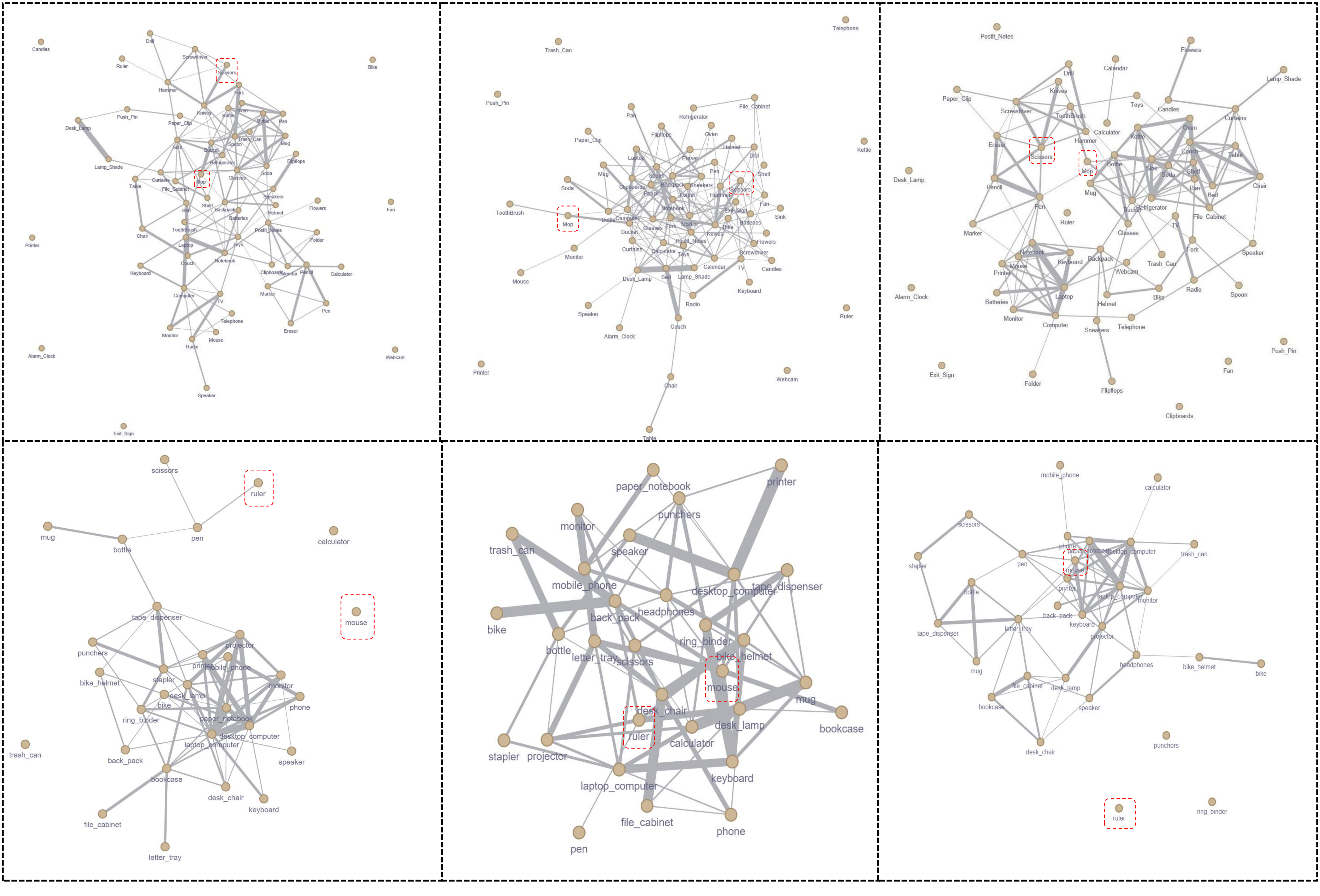}
	\caption{Visualization of three graphs.}
	\label{fig1}
\end{figure*}

\section{Ablation Studies}

We perform ablation studies within the framework of universal domain adaptation using the Office-31 dataset, as detailed in Table~\ref{Table6}. In these experiments, we vary the values of $ \sigma $ in Eq. 7 and Eq. 8 to explore the impact on the sparsity of the adjacency matrix.

As we increased $ \sigma $ to 0.1, we observe a marginal decline in the overall results, approximately around 1\%. Further increments in the value of $ \sigma $ appear to introduce confusion in the model's ability to learn precise relationships.

Additionally, we delve into the impact of various loss functions. For adjacency matrix loss denoted as $\mathcal{L}_{a}$ in Eq. 13, we substitute it with $\mathcal{L}_{1}$ and $\mathcal{L}_{2}$ distance losses, labeling them as UAN + EGLayer w/ $\mathcal{L}_{1}$ and UAN + EGLayer w/ $\mathcal{L}_{2}$, respectively. The corresponding formulations are expressed as follows:

\begin{equation}
\mathcal{L}_{1}(\boldsymbol{A}, \boldsymbol{A}^{s\prime}) = \frac{1}{n^{2}} \sum_{i=1}^{n} \sum_{j=1}^{n} \left(a_{i j} - a^{s\prime}_{i j}\right),
\end{equation}

\begin{equation}
\mathcal{L}_{2}(\boldsymbol{A}, \boldsymbol{A}^{s\prime}) = \frac{1}{n^{2}} \sum_{i=1}^{n} \sum_{j=1}^{n} \left(a_{i j} - a^{s\prime}_{i j}\right)^2.
\end{equation}

We have also employed $L_{1}$ regularization to replace $L_{2}$ regularization in Eq. 14, and we refer to this as UAN + EGLayer w/ $\mathcal{L}_{reg1}$:

\begin{equation}
\mathcal{L}_{reg1}(\boldsymbol{S}^{\prime}) = \| \left\langle \boldsymbol{S}^{\prime}, \boldsymbol{S}^{\prime\mathrm{T}} \right\rangle \|_{1}  = \|\boldsymbol{C}\|_{1} = \sum_{i=1}^n \sum_{j=1}^n | c_{i j} | .
\end{equation}

Moreover, we introduce three versions for ablation studies: UAN + EGLayer w/o $\mathcal{L}_{a}$, UAN + EGLayer w/o $\mathcal{L}_{reg}$, and UAN + EGLayer w/o $\mathcal{L}_{g}$. In the context of these experiments, UAN + EGLayer w/o $\mathcal{L}_{g}$ indicates that the experiment lacks both $\mathcal{L}_{a}$ and $\mathcal{L}_{reg}$.

As indicated in Table~\ref{Table6}, UAN + EGLayer w/ $\mathcal{L}_{1}$ demonstrates a performance improvement when compared to UAN + EGLayer w/o $\mathcal{L}_{a}$. Conversely, UAN + EGLayer w/ $\mathcal{L}_{2}$ shows a performance degradation. Both of these settings underperform in comparison to UAN + EGLayer. Notably, the $L_{1}$ regularization version exhibits a significant performance decrease. This phenomenon could be attributed to regularization causing node embeddings to become too distinct, thereby hindering the model's ability to learn relationships.

Comparing UAN + EGLayer w/o $\mathcal{L}_{a}$, UAN + EGLayer w/o $\mathcal{L}_{reg}$, and UAN + EGLayer w/o $\mathcal{L}_{g}$, it's evident that both $\mathcal{L}_{reg}$ and $\mathcal{L}_{a}$ play distinct roles in boosting performance. When these two losses are not utilized, there is an average performance reduction of 1.06\%. Moreover, the versions with only $\mathcal{L}_{reg}$ and $\mathcal{L}_{a}$ both outperform UAN + EGLayer w/o $\mathcal{L}_{g}$. Finally, UAN + EGLayer demonstrates a clear advantage when compared with UAN + EGLayer w/o $\mathcal{L}_{a}$ and UAN + EGLayer w/o $\mathcal{L}_{reg}$.

Furthermore, we conduct experiments involving 2 GCN layers, with the first layer adapting the features to the same dimension, and the second layer aligning the features to the global graph dimension. This configuration is referred to as UAN + EGLayer + 2 layer GCN. Another setup involve inserting the EGLayer after the feature extractor and before the standard linear classifier as an intermediary layer, without changing dimension. We finally label this experiment as UAN + middle EGLayer.

When compared to the final version, both of these settings exhibit an average performance decrease of 3.63\% and 0.26\%, underscoring the simplicity and effectiveness of the final EGLayer version. It's worth noting that UAN + middle EGLayer demonstrates a 0.61\% and 0.94\% improvements in A→W and W→D domain adaptation, hinting at potential for further exploration when inserting the EGLayer after different layers. Consequently, we remain committed to exploring relevant solutions in this regard.

\begin{table*}[h] 
	\caption{Ablation study for universal domain adaptation experiments on Office-31 dataset}
	\label{Table6}
	\centering
	\begin{tabular}{cccccccc}
	\toprule
	Methods    &  A→W & D→W & W→D & A→D & D→A & W→A & Average \\
	\midrule
UAN + EGLayer + $\sigma$ 0.1 & 83.47 & 93.47 & 93.67 & \bf{86.11} & 86.09 & 85.71 & 88.09 \\ 
UAN + EGLayer + $\sigma$ 0.5 & 15.35 & 28.74 & 24.48 & 20.85 & 21.58 & 11.30 & 20.38 \\ 
UAN + EGLayer & \bf{83.51} & \bf{94.23} & \bf{94.34} & \bf{86.11} & \bf{87.88} & \bf{88.26} & \bf{89.06} \\
\midrule
UAN + EGLayer w/ $\mathcal{L}_{1}$ & 83.89 & 93.58 & 93.57 & 85.69 & 87.38 & 87.24 &  88.56 \\ 
UAN + EGLayer w/ $\mathcal{L}_{2}$ & 83.92 & 93.86 & 93.74 & 85.69 & 84.28 & 85.79 &  87.88 \\ 
UAN + EGLayer w/o $\mathcal{L}_{a}$ & 83.89 & 93.06 & 92.94 & 85.69 & 86.87 & 87.05 & 88.25 \\
UAN + EGLayer w/ $\mathcal{L}_{reg1}$ & 48.23 & 29.86 & 72.45 & 85.71 & 83.56 & 44.84 & 60.78 \\
UAN + EGLayer w/o $\mathcal{L}_{reg}$ & \bf{84.59} & 94.07 & 94.03 & 83.63 & \bf{88.59} & \bf{88.28} & 88.87 \\
UAN + EGLayer w/o $\mathcal{L}_{g}$ & 84.20 & 93.69 & 93.72 & 84.82 & 85.41 & 86.18 & 88.00 \\
UAN + EGLayer & 83.51 & \bf{94.23} & \bf{94.34} & \bf{86.11} & 87.88 & 88.26 & \bf{89.06} \\
\midrule
UAN + EGLayer + 2 layer GCN & 82.79 & 88.71 & 92.79 & 84.31 & 81.72 & 82.27 & 85.43 \\
UAN + middle EGLayer & \bf{84.12} & 93.12 & \bf{95.28} & 85.69 & 87.10 & 87.46 &  88.80 \\
UAN + EGLayer & 83.51 & \bf{94.23} & 94.34 & \bf{86.11} & \bf{87.88} & \bf{88.26} & \bf{89.06} \\
    \bottomrule
	\end{tabular}
\end{table*}

\begin{table*}[h]
	\caption{ViT experiments on miniImageNet and tieredImageNet}
	\label{Tbalevit}
        \setlength\tabcolsep{3pt} 
	\centering
	\begin{tabular}{ccccc}
	\toprule
	dataset & Methods & Backbone & 1-shot & 5-shot \\
	\midrule
    \multirow{7}{*}{miniImageNet} 
    & SUN-NesT \cite{dong2022self} & ViT &  66.54 ± 0.45 &  82.09 ± 0.30 \\  
    & SUN-Visformer \cite{dong2022self} & ViT & 67.80 ± 0.45 & 83.25 ± 0.30  \\
    & FewTURE \cite{hiller2022rethinking} & ViT-S & 68.02 ± 0.88 & 84.51 ± 0.53 \\
    & FewTURE \cite{hiller2022rethinking} & Swin-Tiny & 72.40 ± 0.78 & 86.38 ± 0.49 \\
    & SMKD \cite{lin2023supervised} & ViT-S & 67.98 ± 0.17 &  86.59 ± 0.10 \\
    & SMKD + LPLayer & ViT-S & 73.51 ± 0.19 & 87.15 ± 0.10 \\
    & SMKD + EGLayer & ViT-S & \bf{74.72 ± 0.20} & \bf{88.09 ± 0.10} \\
    \midrule
	\multirow{7}{*}{tieredImageNet}
 
    & SUN-NesT \cite{dong2022self} & ViT &  72.93 ± 0.50 & 86.70 ± 0.33 \\
    & SUN-Visformer \cite{dong2022self} & ViT & 72.99 ± 0.50 & 86.74 ± 0.33 \\
    & FewTURE \cite{hiller2022rethinking} & ViT-S & 72.96 ± 0.92 & 86.43 ± 0.67 \\
    & FewTURE \cite{hiller2022rethinking} & Swin-Tiny & 76.32 ± 0.87 & 89.96 ± 0.55 \\
    & SMKD \cite{lin2023supervised} & ViT-S & \bf{78.50 ± 0.20} & \bf{91.02 ± 0.12} \\
    & SMKD + LPLayer & ViT-S & 78.40 ± 0.20 &  90.96 ± 0.12  \\
    & SMKD + EGLayer & ViT-S & 78.36 ± 0.20 &  90.82 ± 0.12 \\
    \bottomrule
	\end{tabular}
\end{table*}

\begin{table*}[h]
	\caption{Transfer experiments on miniImageNet and tieredImageNet}
	\label{Table7}
        \setlength\tabcolsep{3pt} 
	\centering
	\begin{tabular}{ccccc}
	\toprule
	dataset & Methods & Backbone & 1-shot & 5-shot \\
	\midrule
    \multirow{6}{*}{miniImageNet→tieredImageNet} 
    & Classifier-Baseline & ResNet-12 & 64.15 ± 0.54 & 79.81 ± 0.42 \\
    & Classifier-Baseline + LPLayer & ResNet-12 & 64.51 ± 0.33 & 79.83 ± 0.44 \\
    & Classifier-Baseline + EGLayer & ResNet-12 & \bf{64.89 ± 0.56} & \bf{80.03 ± 0.43} \\
    & Meta-Baseline & ResNet-12 & 67.63 ± 0.49 & 80.99 ± 0.42 \\
    & Meta-Baseline + LPLayer & ResNet-12 & 67.61 ± 0.58 & 80.88 ± 0.43 \\
    & Meta-Baseline + EGLayer & ResNet-12 & \bf{67.98 ± 0.59} & \bf{81.27 ± 0.43} \\
    \midrule
	\multirow{6}{*}{tieredImageNet→miniImageNet}
    & Classifier-Baseline & ResNet-12 & 76.35 ± 0.22 & 90.50 ± 0.12 \\
    & Classifier-Baseline + LPLayer & ResNet-12 & 76.66 ± 0.24 & 90.23 ± 0.12 \\
    & Classifier-Baseline + EGLayer & ResNet-12 & \bf{77.39 ± 0.28} & \bf{91.11 ± 0.14} \\
    & Meta-Baseline & ResNet-12 & 76.79 ± 0.24 & 89.53 ± 0.13 \\
    & Meta-Baseline + LPLayer & ResNet-12 & 78.24 ± 0.29 & 90.41 ± 0.22 \\
    & Meta-Baseline + EGLayer & ResNet-12 & \bf{78.53 ± 0.31} & \bf{90.77 ± 0.13} \\
    \bottomrule
	\end{tabular}
\end{table*}
\begin{table}[h] \footnotesize
	\caption{Zero-shot experiments on miniImageNet and tieredImageNet}
	\label{Table8}
	\centering
	\begin{tabular}{cccc}
	\toprule
	dataset & Methods & 0-shot \\
	\midrule
   miniImageNet & Classifier-Baseline + EGLayer &  48.34 ± 0.20 \\
    tieredImageNet & Classifier-Baseline + EGLayer & 48.50 ± 0.25 \\
    \bottomrule
	\end{tabular}
\end{table}

\section{Implementation Details of Few-Shot Learning}

Our methods are evaluated based on the Matching Networks \cite{vinyals2016matching}, Prototypical Networks \cite{snell2017prototypical}, Classifier-Baseline \cite{chen2021meta}, and Meta-Baseline \cite{chen2021meta}. 

Matching Networks defines a support set $ \mathcal{D}^{S} = \{(\boldsymbol{x}_{i}, \boldsymbol{y}_{i})\} $ and the query images $ \boldsymbol{x}^{\prime} $ for training. The final prediction is calculated as:

\begin{equation}
P(\boldsymbol{y}^{\prime} \mid \boldsymbol{x}^{\prime}, \mathcal{D}^{S})=\sum_{i=1}^k \alpha \left(\boldsymbol{x}^{\prime}, \boldsymbol{x}_{i}\right) \boldsymbol{y}_{i},
\end{equation}
where $ \alpha $ is attention mechanism. 

Prototypical Networks defines prototype $ w_c $ for training by average the embeddings of support set and exploits cosine similarity to calculate the final logits:

\begin{equation}
w_c=\frac{1}{\left|\mathcal{D}^{S}_{c}\right|} \sum_{\boldsymbol{x} \in \mathcal{D}^{S}_{c}} f_\theta(\boldsymbol{x}),
\end{equation}

\begin{equation}
P(\boldsymbol{y}^{\prime}=c \mid \boldsymbol{x}^{\prime}, \mathcal{D}^{S})=softmax(\left\langle f_\theta(\boldsymbol{x}^{\prime}), w_{c}\right\rangle).
\end{equation}

Classifier-Baseline and Meta-Baseline first utilize the whole label-set for training on all base classes with cross-entropy loss. For validation, classifier is removed and feature extractor $ f_{\theta} $ is used to computes the average embedding $ w_c $ of each class $c$ in support set $ \mathcal{D}^{S} $ as Prototypical Networks.

Then, for a query sample $ \boldsymbol{x}^{\prime} $, cosine similarity is computed between the extracted features of $ \boldsymbol{x} $ and average embedding $ w_c $ for the final prediction with softmax function:

\begin{equation}
P(\boldsymbol{y}^{\prime}=c \mid \boldsymbol{x}^{\prime})=\frac{\exp \left(\tau \cdot\left\langle f_\theta(\boldsymbol{x}^{\prime}), w_c\right\rangle\right)}{\sum_{c^{\prime}} \exp \left(\tau \cdot\left\langle f_\theta(\boldsymbol{x}^{\prime}), w_{c^{\prime}}\right\rangle\right)},
\end{equation}
where the Meta-Baseline trains a learnable scalar $ \tau $ through a meta learning way and the Classifier-Baseline fixes the $ \tau $ as 1.0.

\section{Few-shot for Vision Transformers}
We also replace the last linear layer in the prediction head by LPLayer and EGLayer for vision transformers based on prototype version of SMKD \cite{lin2023supervised}, called SMKD + LPLayer and SMKD + EGLayer, respectively. As shown in Table~\ref{Tbalevit}, we observe that SMKD + LPLayer holds a clear advantage in the 1-shot settings, achieving improvements of 4.53\% compared with baseline. Furthermore, SMKD + EGLayer demonstrates an additional 1.21\% improvement compared to SMKD + LPLayer. Under 5-shot setting, SMKD + LPLayer shows a modest promotion of 0.56\%, while SMKD + EGLayer enhances the baseline by 1.50\% on miniImageNet. For tieredImageNet, despite a marginal decline in the performance of both SMKD + LPLayer and SMKD + EGLayer compared to the baseline, their overall performance remains stable. In summary, these experiments demonstrate that EGLayer is not only easily adaptable in ConvNet and ResNet but also functions as a suitable plug-and-play module for Vision Transformers (ViTs), showcasing its potential efficacy in large vision models. In the future, we plan to explore additional ways for EGLayer adapting to ViTs to get better performance, including the design of optimal insertion points for different layers and loss functions.

\section{Transfer Learning}
We have evaluated the transfer learning ability of our method by exchanging the models trained on miniImageNet and tieredImageNet in both Classifier-Baseline and Meta-Baseline settings. We name the model trained on miniImageNet for few-shot learning on tieredImageNet as miniImageNet→tieredImageNet, and vice versa. As shown in Table \ref{Table7}, Meta-Baseline + EGLayer achieves the best performance for both 1-shot (67.98\%) and 5-shot (81.27\%) in miniImageNet→tieredImageNet setting. 

In the tieredImageNet→miniImageNet setting, Classifier-Baseline + EGLayer outperforms Classifier-Baseline and Classifier-Baseline + LPLayer, improving the performance by 1.04\%$/$0.61\% and 0.73\%$/$0.88\%, respectively. For Meta-Baseline, Meta-Baseline + EGLayer still have 1.74\%$/$1.24\% and 0.29\%$/$0.36\% improvements over Meta-Baseline and Meta-Baseline + LPLayer. In general, our method demonstrates an overall advantage in transfer learning tasks, validating the generalization and reliability of the learned features.

\section{Zero-Shot Learning}

We conduct zero-shot experiments to evaluate whether EGLayer could align extracted features more closely with the semantic space by leveraging external knowledge. In these experiments, we employ graph node embeddings instead of one-shot image features, facilitating zero-shot learning. The results presented in Table \ref{Table8} indicate that EGLayer achieved an accuracy of approximately 50\% in zero-shot tasks. This result confirms the ability of EGLayer to align features with the semantic space effectively.

\end{document}